\documentclass[conference]{IEEEtran}

\usepackage{times}
\usepackage{epsfig}
\usepackage{graphicx}
\usepackage{amsmath}
\usepackage{amssymb}
\usepackage{amsmath,graphicx,amssymb,multirow}
\usepackage{subfig}
\usepackage{epsfig}
\usepackage{tabularx,booktabs}

  \newcolumntype{C}{>{\centering\arraybackslash}X} % centered version of "X" type

\begin{document}

%%%%%%%%% TITLE
\title{Generating High Quality Visible Images from SAR Images Using CNNs}

% Authors at the same institution
%\author{First Author \hspace{2cm} Second Author \\
%Institution1\\
%{\tt\small firstauthor@i1.org}
%}
% Authors at different institutions
\author{Puyang Wang \hspace{2cm} Vishal M. Patel \\
Department of Electrical and Computer Engineering\\
Rutgers, The State University of New Jersey\\
94 Brett Rd, Piscataway, NJ 08854\\
{\tt\small \{puyang.wang, vishal.m.patel\}@rutgers.edu}
}

\maketitle

%%%%%%%%% ABSTRACT
\begin{abstract}
We propose a novel approach for generating high quality visible-like images from Synthetic Aperture Radar (SAR) images using Deep Convolutional Generative Adversarial Network (GAN) architectures.  The proposed approach is based on a cascaded network of  convolutional neural nets (CNNs) for despeckling and image colorization.  The cascaded structure results in faster convergence during training and produces high quality visible images from the corresponding SAR images.   Experimental results on both simulated and real SAR images show that the proposed method can produce visible-like images better compared to the recent state-of-the-art deep learning-based methods. 
\end{abstract}


%%%%%%%%% BODY TEXT
\section{Introduction}

Synthetic aperture radar (SAR) is a coherent radar imaging technology which is capable of producing high-resolution images of targets and landscapes.  Due to its ability to capture images both at night and in bad weather conditions, SAR imaging has several advantages compared to optical and infrared systems.
However, SAR images are often difficult to interpret mainly due to the following two reasons. 
\begin{enumerate}
\item They are contaminated by multiplicative noise known as speckle.   Speckle is caused by the constructive and destructive interference of the coherent returns scattered by small reflectors within each resolution cell \cite{Speckle_Goodman}.   
\item  Processed SAR images are often grayscale and they do not contain any color information.  
\end{enumerate}
 These two issues often make the processing and interpretation of SAR images very difficult for both human interpreters and computer vision systems.     Hence, despeckling and proper colorization are important for semantically interpreting the reflectivity field in SAR imaging.  

Assuming that the SAR image is an average of $L$ looks, the observed SAR image $Y$ is related to the noise free image $X$ by the following multiplicative model \cite{Book_Ulaby}
\begin{equation}\label{eq:multiplicative}
	Y = F\odot X,
\end{equation}
where $F$ is the normalized fading speckle noise random variable and $\odot$ denotes the element-wise multiplication.  One common assumption on $F$ is that it follows a Gamma distribution with unit mean and variance $\frac{1}{L}$ and has the following probability density function \cite{noisemodel}
\begin{align} \label{eq:pdf}
	\centering
	p(F) = \frac{1}{\Gamma(L)}L^LF^{L-1}e^{-LF},
\end{align}
where $\Gamma(\cdot)$ denotes the Gamma function and $F \geq 0$, $L \geq 1$.

\begin{figure}[t]
 \centering \includegraphics[width=.49\linewidth]{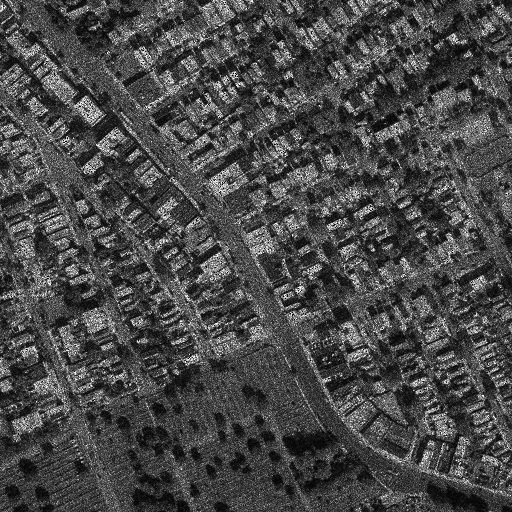} \includegraphics[width=0.49\linewidth]{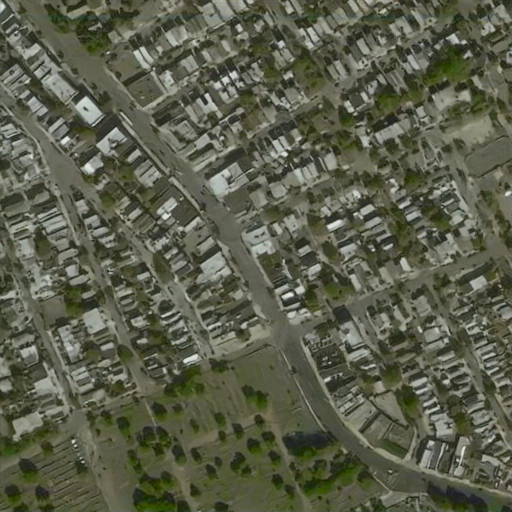}\\
 (a)\hskip120pt (b)
\caption{A sample result of the proposed SAR-GAN method for SAR image to visible image translation. (a) Simulated input noisy SAR image. (b) Despeckled and colorized image.}
\label{fig:sample_results} 
\end{figure}

Based on the above SAR observation model, various methods have been developed in the literature to suppress speckle.  These include multi-look processing \cite{Book_SAR_IMU, Thompson_SAR}, filtering methods \cite{lee1981speckle, frost, gammamap}, wavelet-based despecking methods \cite{DeSpeckle_wavelet_MRF, Despeckle_Wavelet_undecimeted, Despeckle_wavelet_heavytail, Despeckle_MCA}, SAR block-matching 3D (SAR-BM3D) algorithm \cite{sarbm3d}, Total Variation (TV) methods \cite{Despeckle_TV}, and deep learning-based methods \cite{ID-CNN, Puyang_CAMSAP}.   
Note that some of these methods apply homomorphic processing in which the multiplicative noise is transformed into an additive noise by taking the logarithm of the observed data \cite{SAR_Despeckle_Tutorial}.  

Although state-of-the-art SAR image despecking algorithms such as SAR-BM3D and wavelet-based methods are able to generate despeckled SAR images with sharp edges, the resulting despeckled images are often difficult to interpret due to their grayscale nature.  For example, even after despecking, it is difficult to distinguish between  sandy land and grass field due to the grayscale nature of SAR images.  Hence, generating a visible-like image from a SAR image is not only an interesting problem but also important for semantic segmentation and interpretation of SAR images. This problem shares some similarities with image colorization. However, there are some notable differences. First, in the image colorization domain (grayscale image to RGB) the luminance is directly given by grayscale input, so only the chrominance needs to be estimated. 
Secondly, in the case of colorization techniques, in general noiseless grayscale images are given as input to
obtain the RGB images. But in the case of SAR images, the input will have speckle and the expected output is the clean visible-like image with three RGB channels.

In this paper, we develop a deep learning-based method, called SAR-GAN, for the problem of SAR image to high quality visible image translation where we map a single channel noisy SAR image into a visible-like RGB image.   Figure~\ref{fig:sample_results} shows a sample output from our SAR-GAN method.  Given a simulated speckled SAR image shown in Figure~\ref{fig:sample_results} (a), SAR-GAN can generate not only the despeckled image but also the visible-like image as shown in  Figure~\ref{fig:sample_results} (b).   As can be seen by comparing Figure~\ref{fig:sample_results} (a) and Figure~\ref{fig:sample_results} (b), that our method is able to simultaneously denoise and colorize the simulated SAR image reasonably well.

\section{Proposed Method}
In this section, we provide details of the proposed SAR-GAN method in which we aim to learn a mapping from input speckled SAR images to visible images for both noise removal and colorization.  The proposed method consists of  three main components: despeckling sub-network $G_{D}$, colorization sub-network $G_{C}$ and generative adversarial learning. The primary goal of the despeckling sub-network is to restore a clean image from a noisy observation. The colorization  sub-network then transforms the despeckled image into a visible image. Inspired by recent works on using generative adversarial learning for image colorization, we add the adversarial loss by introducing a discriminator network $D$.   The adversarial loss, empirically, can in principle become aware that gray looking outputs are unrealistic, and encourage a wider color distribution. 
The composition of the two sub-networks, despeckling and colorization, forms the generator $G$ in a typical generative adversarial network (GAN) framework as follows:
\begin{equation}
G = G_C \circ G_D.
\end{equation}

\begin{figure}[htp!]
	\centering
	\includegraphics[width=90mm]{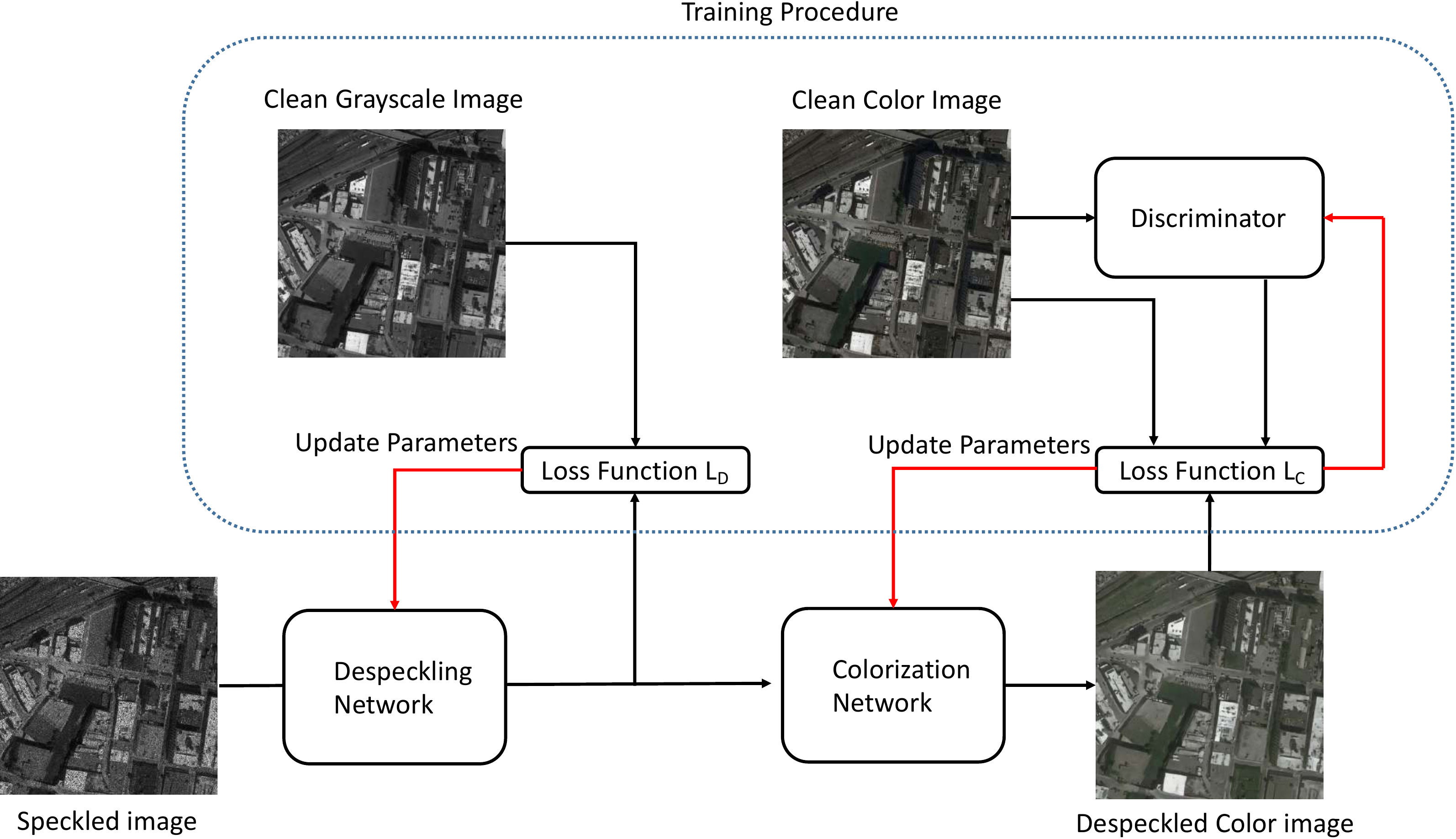}
	\caption{Proposed SAR-GAN network architecture for SAR to visible image translation.}
	\label{fig:network}
\end{figure}

The overall structure of the proposed SAR-GAN method containing two sub-networks and the training procedure is shown in Figure~\ref{fig:network}, where black arrow lines indicate data flows and red arrow lines denote network parameter updating.  The detailed architectures of both sub-networks and loss functions are discussed in the following subsections.

\subsection{Despeckling Network}
\begin{figure}[htp!]
	\centering
	\includegraphics[width=90mm]{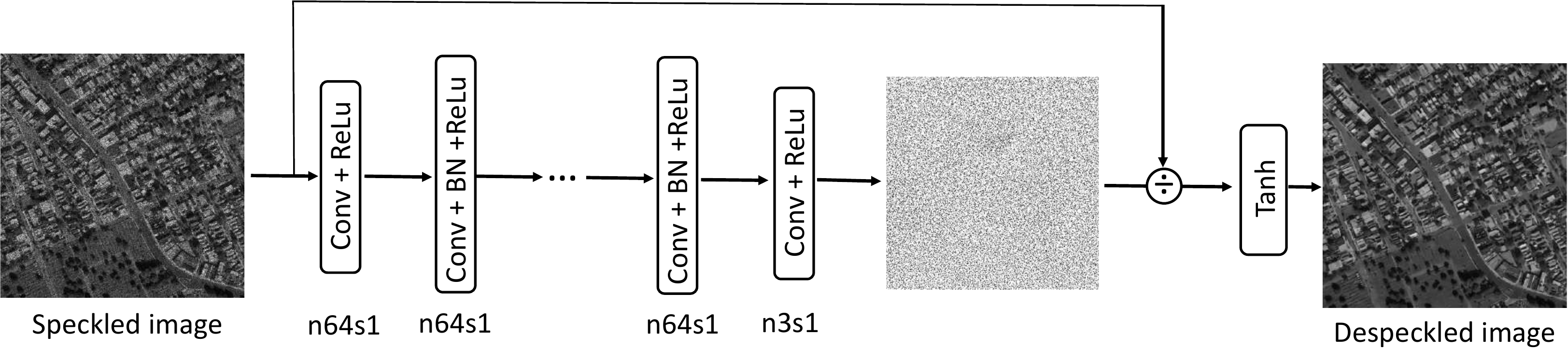}
	\caption{Proposed network architecture for image despeckling.}
	\label{fig:ds_network}
\end{figure}

 The detailed architecture of the despeckling sub-network is shown in Figure~\ref{fig:ds_network}, where Conv, BN and ReLu stand for Convolution and Batch Normalization and Rectified Linear Unit, respectively.  The despeckling CNN is adopted from our previous work \cite{ID-CNN} on SAR image restoration. Using a specific CNN architecture, we learn a mapping from an input SAR image into a despeckled image.  One possible solution to the despeckling problem would be to transform the image into a logarithm space and then learn the corresponding mapping via CNN \cite{sarcnn}. However, this approach needs extra steps to transfer the image into a logarithm space and from a logarithm space back to an image space. As a result, the overall algorithm can not be learned in an end-to-end fashion. To address this issue, a division residual method is leveraged in our method where a noisy SAR image is viewed as a product of speckle with the underlying clean image (i.e. \eqref{eq:multiplicative}).  By incorporating the proposed component-wise division residual layer into the network,  the convolutional layers are forced to learn the speckle component during the training process.  In other words, the output before the division residual layer represents the estimated speckle.  Then, the despeckled image is obtained by simply dividing the input image by the estimated speckle.  

The noise-estimating part of despeckling sub-network consists of 8 convolutional layers (along with batch normalization and ReLU activation functions), with appropriate  zero-padding to make sure that the output of each layer shares the same dimension with that of the input image. Batch normalization is added to alleviate the internal covariate shift by incorporating a normalization step and a scale and shift step before the nonlinearity in each layer. Each convolutional layer (except for the last convolutional layer) consists of 64 filters with stride of one. Then, the division residual layer with skip connection divides the input image by the estimated speckle. A hyperbolic tangent layer is stacked at the end of the network which serves as a non-linear function. 

%{\bf{It seems like you use our SPL network here.  We need to cite it.}}

\subsection{Colorization Network}
\begin{figure}[htp!]
	\centering
	\includegraphics[width=90mm]{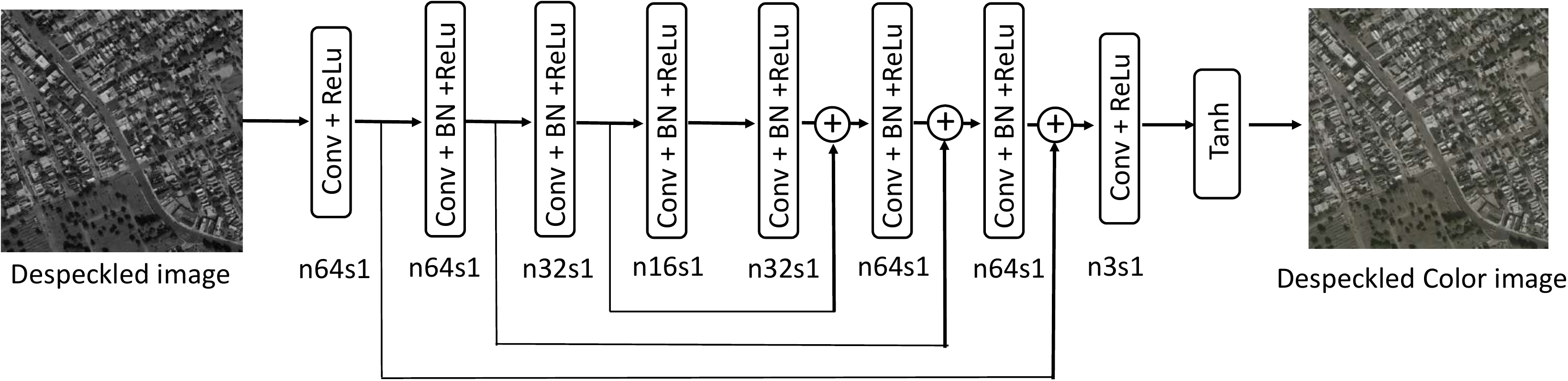}
	\caption{Proposed network architecture for image colorization.}
	\label{fig:c_network}
\end{figure}

Deep learning-based image colorization has been studied over the last couple of years \cite{colorization_zhang}, \cite{colorization_lizuka}. The key part of an image colorization neural network is to fully leverage the contextual information of an image for color translation. To extract and utilize the contextual information, one common way in deep learning is to use an encoder-decoder architecture in which an input image is encoded into a set of feature maps in the middle of the network. However, such a network requires that all information flow passes through all the layers.  For the image colorization problem, the sharing of low-level information between the input and output is important since the input and output should share the location of prominent edges. For the above reason, we add skip connections, following the general shape of an encoder-decoder CNN \cite{skipconnection} as shown in Figure~\ref{fig:c_network}. 

The colorization sub-network forms a symmetric encoder-decoder with 8 convolution layers and 3 skip connections. For each convolution layer, the kernel size is $3 \times 3$. Note that the $n64s1$ in Figure~\ref{fig:ds_network} and \ref{fig:c_network} stands for 64 feature maps with one stride.

\subsection{Loss Functions}
In a SAR image translation problem, it is important that the output image is noise free and realistic. One common loss function used in many image translation problems is the $L_{1}$ loss. Given an image pair $(X, Y)$, where $Y$ is the noisy input image and $X$ is the corresponding ground truth, the per-pixel $L_{1}$ loss is defined as
\begin{equation}
\nonumber L_{L_1}(Y,X;G) = \frac{1}{CWH}\sum_{w=1}^{W}\sum_{h=1}^{H}\|G(Y^{c,w,h})-X^{c,w,h}\|_1,
\end{equation}
where $G$ is the learned network and $G(Y^{c,w,h})$ is the filtered image. 
Note that we have assumed that $X$ and $Y$ are of the same size $C\times W \times H$ where $C$ stands for the number of color channels.
In this case, the network is trained to minimize the $L_{1}$ distance between the output and the ground truth on the training set. Although the $L_{1}$ loss has been shown to be very effective for image de-noising problem, it will incentivize an average, grayish color when it is uncertain which of several plausible color values a pixel should take on. In particular, $L_{1}$ will be minimized by choosing the median of the conditional probability density function over possible colors. Hence, the $L_{1}$ loss alone is not suitable for image colorization. Recent studies have shown adversarial loss, on the other hand, can in principle become aware that gray looking outputs are unrealistic, and encourage matching the true color distribution. 

Given a set of $N$ despeckled and colorized images,  $\{\hat{X}^{c,w,h}_{i}\}_{i=1}^{N}$, generated from the generator $G$, the adversarial loss to guide the generator is defined as 
\begin{equation}
L_A(\hat{X}; D, G) = -\frac{1}{N}\sum_{i=1}^{N}\log(D(\hat{X}^{c,w,h}_i)),
\end{equation}
where $\hat{X}^{c,w,h}_{i}=G(Y^{c,w,h}_{i})$.  
One of the issues with the adversarial loss is that it does not rely on the ground truth $X$.   Hence, the results often contain artifacts that are not present in the clean ground truth image.  It tries to make the 'style' of the output closer to the training images.

Considering the pros and cons of both losses, we combine the per-pixel $L_{1}$ loss and the adversarial loss together with appropriate weights to form our new refined loss function. The proposed loss function is defined as follows
\begin{align} \label{eq:ds_loss}
&L_{D} = L_{L1}(gray(Y), gray(X); G_{D}),\\
\label{eq:c_loss}
&L_{C} = L_{L1}(Y, X ;G_{C})+ \lambda_a L_A(\hat{X}; D, G_{C}),\\
\label{eq:loss}
&L = L_{D} + L_{C},
\end{align}
where $gray(X)$ and $gray(Y)$ are the corresponding grayscale versions of ground truth $X$ and noisy $Y$ with single channel, respectively.  Here, $L_{D}$ and $L_{C}$ are loss functions for despeckling and colorization sub-network, respectively. The overall function $L$ is the sum of $L_{D}$ and $L_{C}$. The $L_1$ loss in \eqref{eq:ds_loss} makes the despeckling network $G_D$ learn a mapping between the speckled input and clean ground truth. Loss function $L_C$ for the colorization sub-network is a weighted sum of  $L_1$ and adversarial loss. Note that, for SAR images, the number of color channels is equal to 1. Hence, the dimension of input $Y$ and $G_{D}(Y)$ should be $1\times W \times H$ and $3\times W \times H$ for $X$ and $G(Y)$. $\lambda_a$ is a pre-defined weight for adversarial loss to balance the scale difference between losses. Because of the single combined loss function $L$ we are able to train the network $G$ which contains two sub-networks in an end-to-end fashion.

\section{Experimental Results}
To evaluate the effectiveness and performance of our proposed method, we present and compare results of our SAR-GAN with others methods. Since no similar work on despeckling and colorization of SAR images simultaneously has been done, we compare the performance of our method with that of the
two CNN methods (CNN \cite{cnn} and pix2pix \cite{pix2pix}) and their combinations with the state-of-the-art despeckling algorithm SAR-BM3D \cite{sarbm3d}.  For all the compared methods, parameters are set as suggested in their corresponding papers.  
For the basic CNN method, we adopt the network structure proposed in \cite{cnn} and train the  network using the same training dataset as used  to train our network.   

To train the proposed SAR-GAN network, we generate a dataset that contains 3292 image pairs. Training images are collected from the scraped Google Maps images \cite{pix2pix} and the corresponding speckled images are generated using \eqref{eq:multiplicative}.  All images are resized to $512 \times 512$. The entire network is trained  using the ADAM optimization method  \cite{adam_opt}, with mini-batches of size 12 and learning rate of 0.0002. During training, we set $\lambda_a = 0.1$.The architecture of the discriminator $D$ is adapted from that in \cite{GAN_D}.

\begin{figure*}[htp!]
	\centering
	\includegraphics[width=50mm]{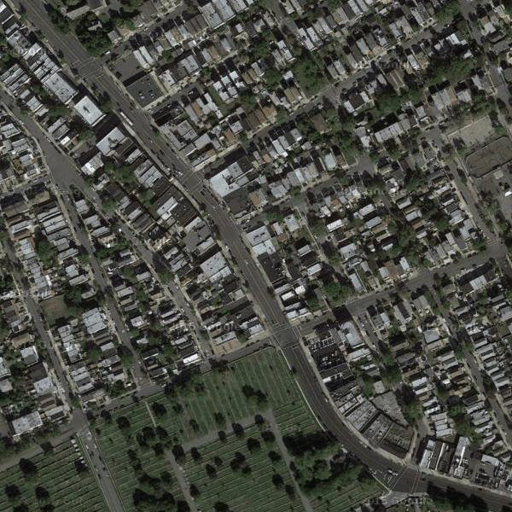}
	\includegraphics[width=50mm]{figs/noisy.png}
	\includegraphics[width=50mm]{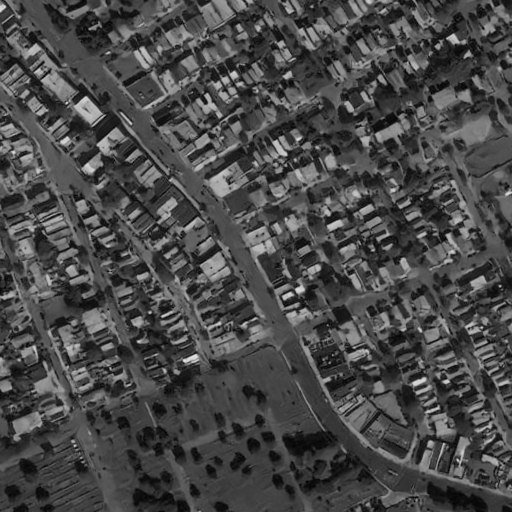}\\
	(a) \hspace{13em}(b)\hspace{13em}(c)\\
	\vspace{0.5em}
	\includegraphics[width=50mm]{figs/vis_our.png}
	\includegraphics[width=50mm]{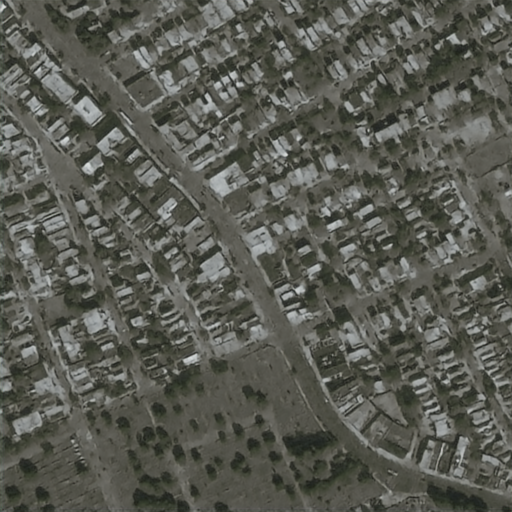}
	\includegraphics[width=50mm]{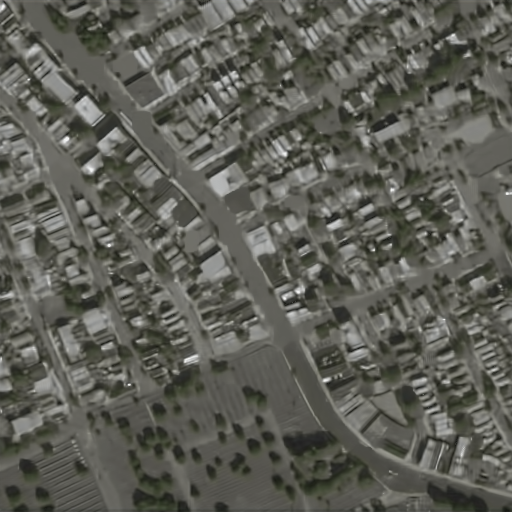}\\
	(d) \hspace{13em}(e)\hspace{13em}(f)\\
	\vspace{0.5em}
	\includegraphics[width=50mm]{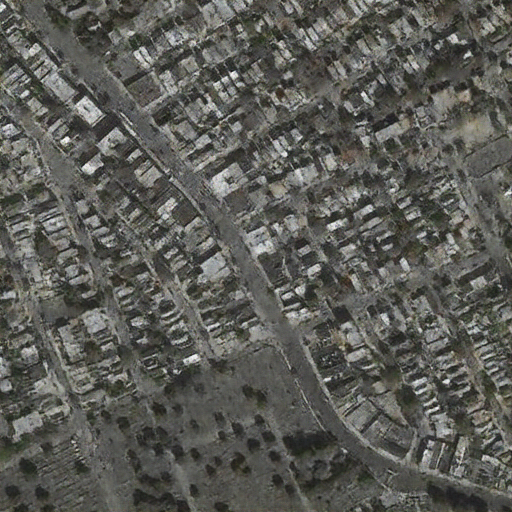}
	\includegraphics[width=50mm]{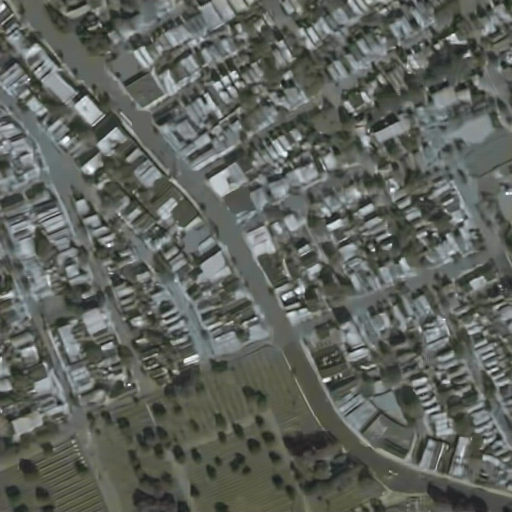}\\
	(g) \hspace{13em}(h)
	\caption{(a) Ground truth. (b) Synthetic SAR image. (c) SAR-GAN despeckled. (d) SAR-GAN. (e) CNN. (f) CNN w/ despeckling. (g) pix2pix. (h) pix2pix w/ despeckling. }
	\label{fig:syn}
\end{figure*}

 \begin{figure*}[htp!]
	\centering
	\includegraphics[width=82mm]{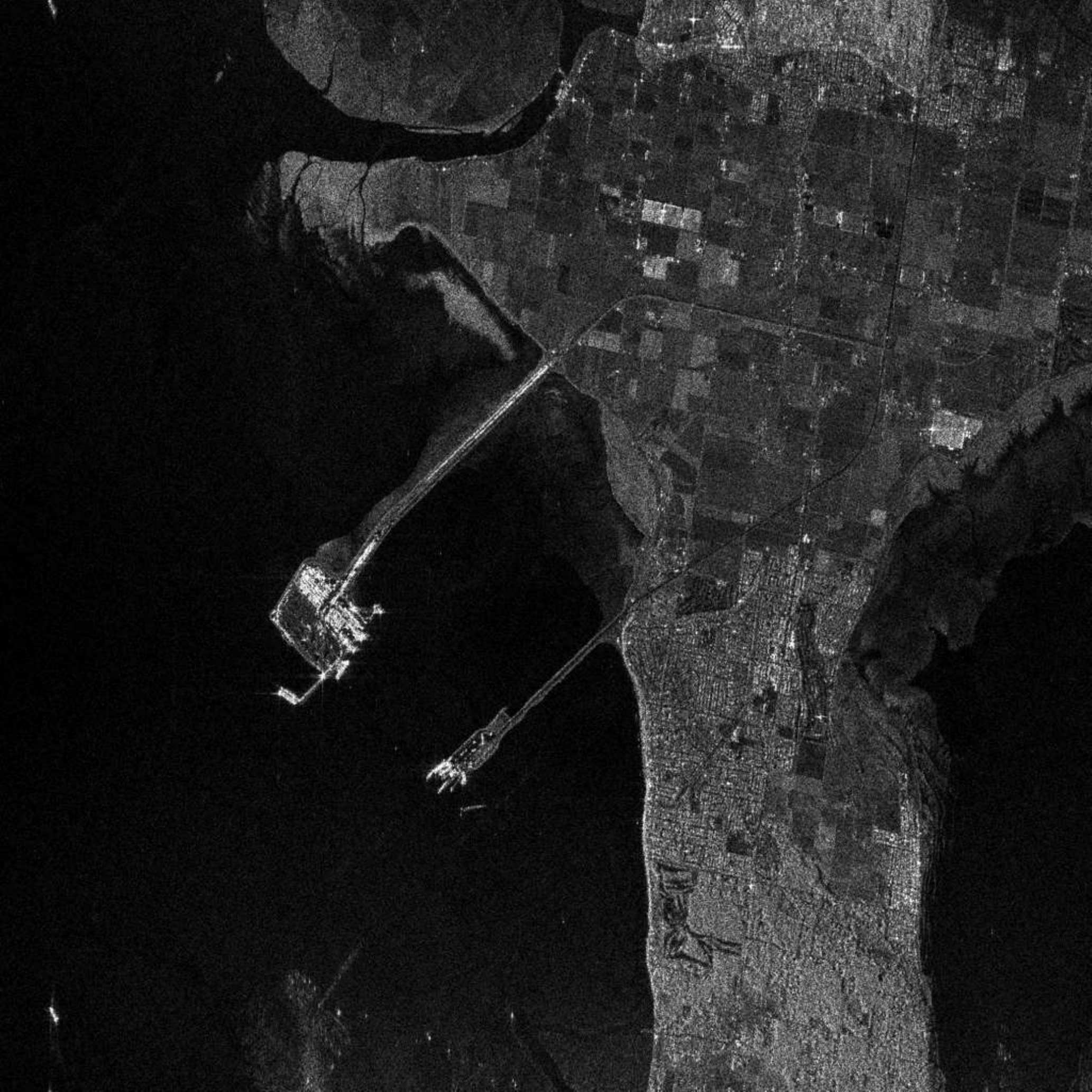}
	\includegraphics[width=82mm]{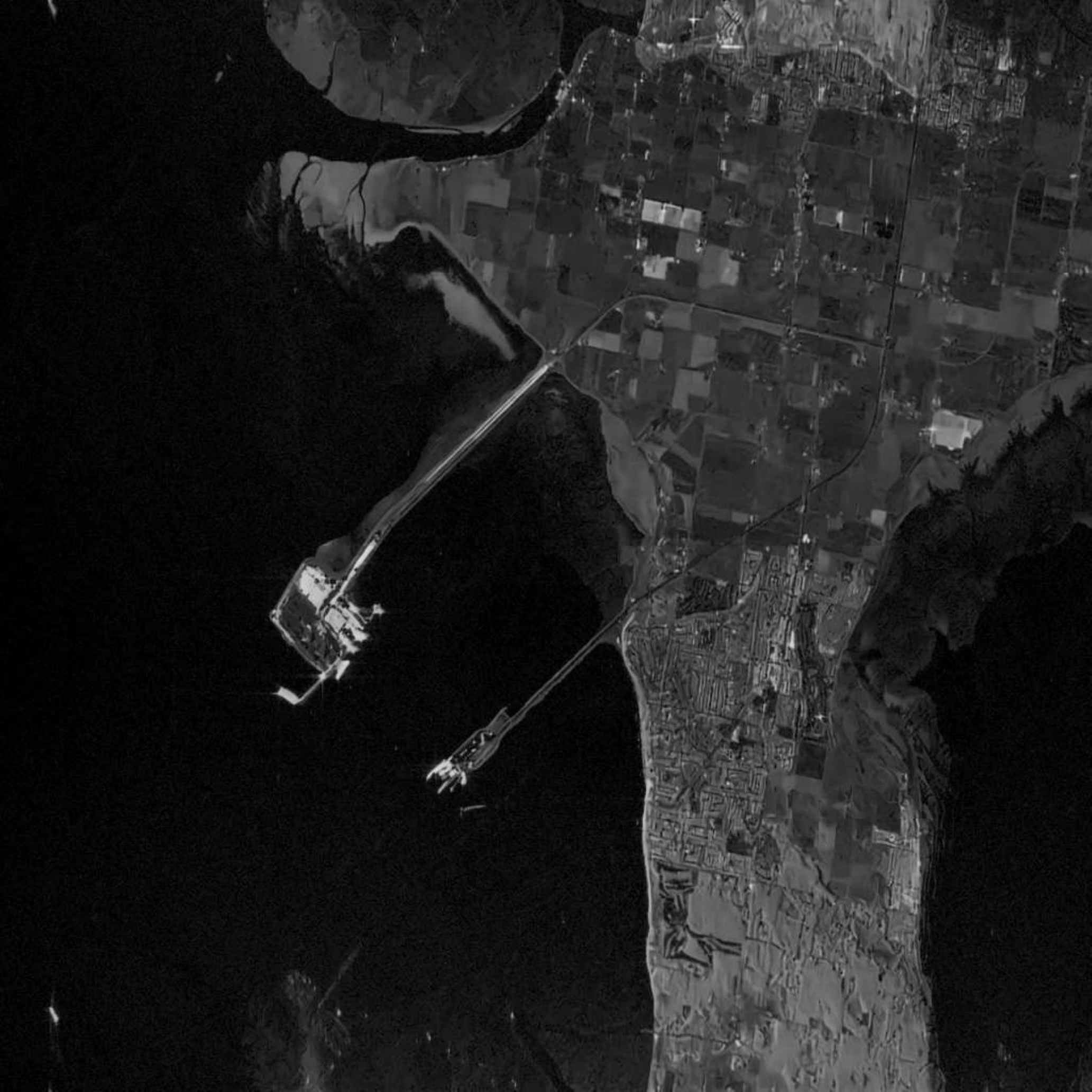}\\
	\hspace{5em}(a) SAR image. \hspace{13em} (b) despeckled image by SAR-GAN.\\
	\vspace{0.5em}
	\includegraphics[width=82mm]{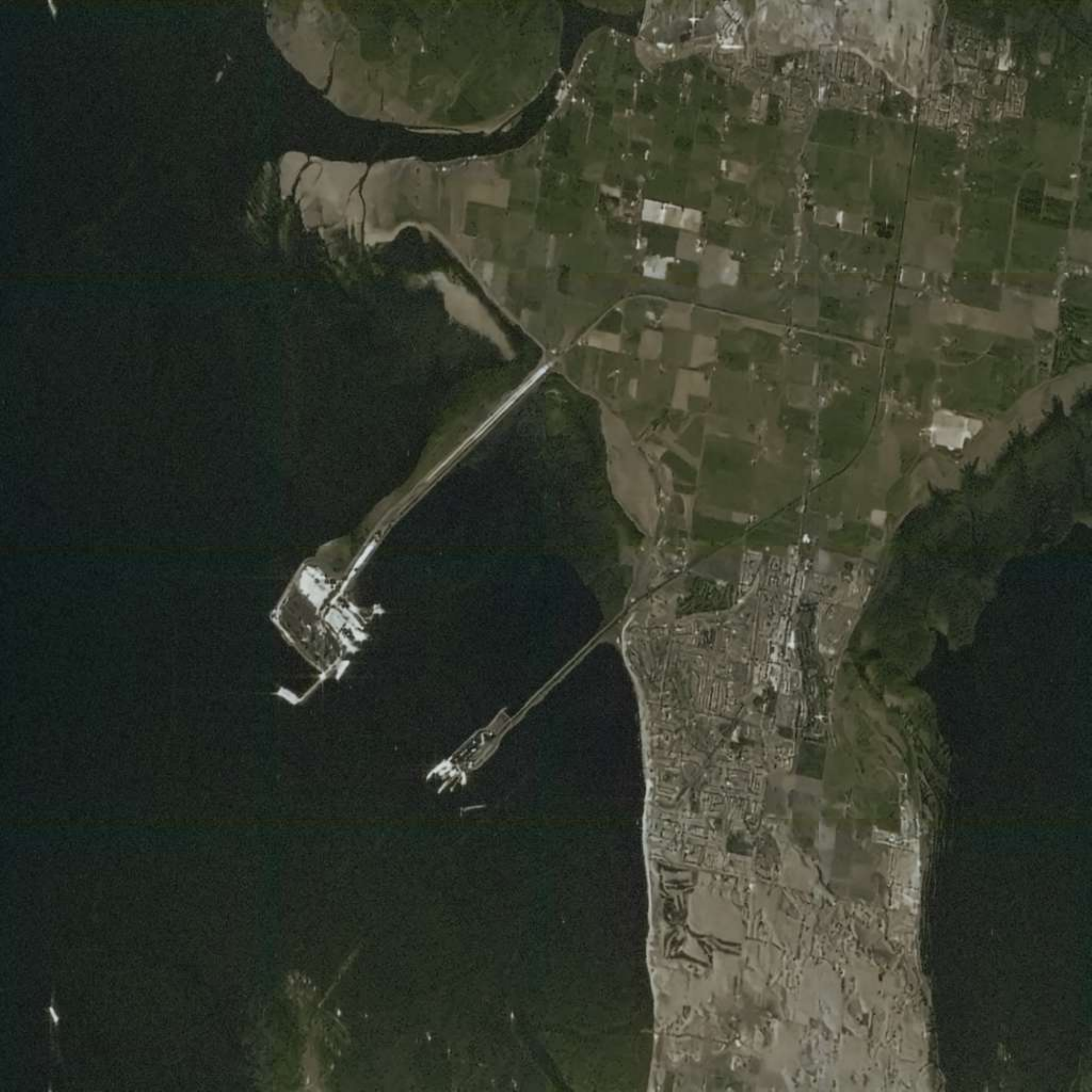}
	\includegraphics[width=82mm]{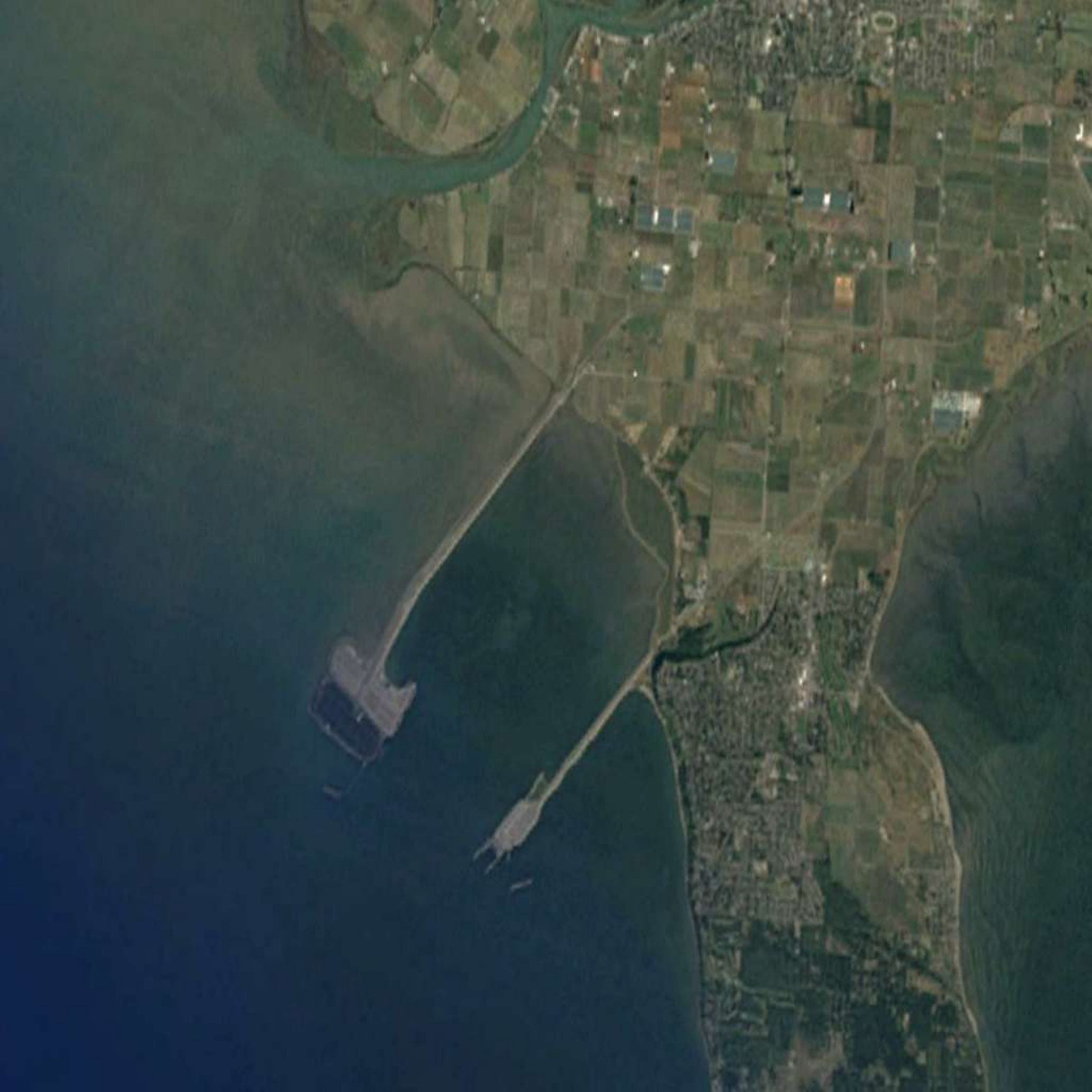}\\
	(c) Visible image by SAR-GAN. \hspace{13em} (d) Satellite image.\\
	\caption{Results of SAR-GAN on a real SAR image.}
	\label{fig:real}
\end{figure*}

\subsection{Despeckling Performance}
One key part of generating high quality visible images from SAR images is about removing as much speckle as possible meanwhile retaining the fine details. Therefore, we perform an experiment comparing despeckling performance of the proposed SAR-GAN and other SAR image despeckling algorithms including the state-of-the-art SAR-BM3D on synthetic SAR images. The outputs of the despeckling network $G_D$ are used for comparison.

We randomly selected 85 speckled images out of all images in the dataset.  The remaining images are used for training the network.  Experiments are carried out on three different noise levels. In particular, the number of looks $L$ is set equal to be 1, 4 and 10, respectively.  The Peak Signal to Noise Ratio (PSNR), Structural Similarity Index (SSIM) \cite{ssim}, Universal Quality Index (UQI) \cite{UQI}, and Despeckling Gain (DG) \cite{benchmark} are used to measure the denoising performance of different methods.  Average results of 85 test images corresponding this experiment are shown in Table~\ref{tab:synthetic-result}.  As can be seen from this table, in all three noise levels, SAR-GAN provides the best performance compared to the other despeckling methods.

\begin{table}[htp!]
	\renewcommand{\arraystretch}{1}
	\caption{Quantitative results for various experiments on synthetic images.}
	\label{tab:synthetic-result}
	\centering
	\resizebox{\linewidth}{!}{%
		\begin{tabular}{c|c|c|c|c|c|c|c}
			\hline
			\hline
			&Metric  &  Noisy & Lee & Kuan & PPB & SAR-BM3D & ID-CNN \\
			\hline
			
			&PSNR   &  14.53 & 21.48 & 21.95 & 21.74 & 22.99 &  \textbf{24.74}\\
			
			$L=1$&SSIM    &  0.369 & 0.511 & 0.592 & 0.619 & 0.692 &  \textbf{0.727}\\
			
			&UQI       & 0.374 & 0.450 & 0.543 & 0.488 & 0.591 &  \textbf{0.621}\\
			
			&DG        & - & 16.01 & 17.08 & 14.30 & 17.17 &  \textbf{23.51}\\
			\hline
			&PSNR   &  18.49 & 22.12 & 22.84 & 23.72 & 24.96 &  \textbf{26.89}\\
			
			$L=4$&SSIM& 0.525 & 0.555 & 0.650 & 0.725 & 0.782 &  \textbf{0.818}\\
			
			&UQI    & 0.527 & 0.485 & 0.594 & 0.605 & 0.679 &  \textbf{0.723}\\
			
			&DG        & - & 8.35 & 10.00 & 10.52 & 14.89 &  \textbf{19.33}\\
			\hline
			&PSNR   &  20.54 & 22.30 & 23.11 & 24.92 & 26.45 &  \textbf{28.07}\\
			
			$L=10$&SSIM&0.602 & 0.571 & 0.671 & 0.779 & 0.834 &  \textbf{0.853}\\
			
			&UQI    & 0.599 & 0.498 & 0.613 & 0.678 & 0.745 &  \textbf{0.765}\\
			
			&DG     & - & 4.06 & 5.93 & 7.75 & 13.61 &  \textbf{17.35}\\
			\hline
			\hline
	\end{tabular}}
\end{table}

\subsection{Results on Synthetic Images}
The despecking as well as colorization results on a synthetic image corresponding to different methods are shown in Figure~\ref{fig:syn}.  The details of the four compared methods are as follows:
\begin{itemize}
	\item \textbf{CNN} Network is adopted from \cite{cnn} and trained with only the $L_1$ loss. The input and output are speckled image and generated visible image, respectively.
	\item \textbf{CNN w/ SAR-BM3D} The input images are first despeckled by SAR-BM3D and then fed into the network which is trained on image pairs of grayscale image and the corresponding color image.
	\item \textbf{pix2pix} The $L_1$+cGAN network in \cite{pix2pix} trained with the $L_1$ and the adversarial losses. The input and output are speckled image and generated visible image, respectively.
	\item \textbf{pix2pix w/ SAR-BM3D} Similar to CNN w/ SAR-BM3D, the colorization network is replaced by the pix2pix network.
\end{itemize}

From Figure~\ref{fig:syn}, we can clearly see that our proposed SAR-GAN performs the best overall. Compared with Figure~\ref{fig:syn} (e) and (g), our result in Figure~\ref{fig:syn} (d) suffers from less artifacts because of better despeckling performance of the despeckling network.   Furthermore, from (f) and (h) we see that  SAR-BM3D helps to suppress speckle but at the cost of losing some detail information. Note that (e) and (f) both have some gray color in the final output.  We believe that this is mainly due to the use of only the $L_1$ loss in their networks.

\subsection{Results on Real SAR Images}
Finally, we evaluate the performance of the proposed SAR-GAN on a real SAR image. Results are shown in Figure~\ref{fig:real}. The real SAR image shown in Figure~\ref{fig:real} (a) is from the Vancouver scene of RADARSAT-1 operating on the C band \cite{Book_Cumming}. Parameters of RADATSAT-1 for the Vancouver scene are as follows:\\
Sampling Rate $F_r$ is 32.317 MHz, pulse duration $T_r$ is 41.7 $\mu$s and radar frequency $f_0$ is 5.3 GHz.   
Figure~\ref{fig:real} (d) is the satellite image captured on the same date as in (a). By comparing Figure~\ref{fig:real} (c) and (d), we can clearly see that the proposed SAR-GAN is capable of generating high quality visible-like image from a real SAR image.

\section{Conclusion}
We proposed a novel approach for generating high quality visible-like images from SAR images using GAN architectures. The proposed approach is based on the usage of a cascaded model for despeckling and colorization in a progressive way. The cascaded structure allows a fast convergence during the training and obtains a greater similarity between the given SAR image and the corresponding visible image. The proposed approach has been evaluated on both simulated and real SAR images and it is shown that the proposed  approach can provide better colorization compared to some of the recent deep learning-based methods.  
\section*{Acknowledgment}
This work was supported by an ARO grant W911NF-16-1-0126.
\bibliographystyle{IEEEtran}
\bibliography{sar}

\end{document}